# Computationally-Optimal Real-Resource Strategies for Independent, Uninterruptible Methods

David Einav*

Rockwell International Science Center, Palo Alto Laboratory,
444 High Street Palo Alto, CA 94301

and

Department of Engineering-Economic Systems
Stanford University, Stanford, CA 94305

Michael R. Fehling[†]

Department of Engineering-Economic Systems
Stanford University, Stanford, CA 94305

## Abstract

This paper focuses on managing the cost of deliberation before action. In many problems, the cost and the resource consumption of the deliberation phase cannot be ignored, and the overall quality of the solution reflects the costs incurred and the resources consumed in deliberation as well as the cost and benefit of execution. A feasible strategy that minimizes the total cost is termed *computationally-optimal*. For a situation where a number of independent, uninterruptible methods are available to solve the problem, we develop a pseudopolynomial-time algorithm to construct generate-and-test computationally-optimal strategies. Stochastic Dynamic Programming is used to solve the problem that is shown to be NP-complete and the results address problems occurring in automatic emergency-response systems, design automation, query optimization, destructive testing, and other areas characterized by significant computational costs or limited deliberation resources.

## 1 Introduction

This paper focuses on managing the cost of deliberation before action. In many problems the cost and the resource consumption of the deliberation phase cannot be ignored, and the overall quality of the solution reflects the costs incurred and the resources consumed in deliberation as well as the cost and benefit of execution.

We consider the situation where a number of independent, uninterruptible methods are available to solve the problem. The methods are characterized by uncertain cost and resource consumption, and are sequentially selected and evaluated. The selection process is performed by a strategy that determines the next method to be evaluated based on the methods selected so far as well as the results of previous evaluations. After a sequence of methods is computed, the strategy halts the process at which point the best solution found so far is executed.

Our goal is essentially to formalize the tradeoff between the costs of deliberation and the benefit of immediate action by developing a family of algorithms to construct generate-and-test strategies that are optimal with respect to expected global cost and have limited resource consumption in the deliberation phase. Such a feasible strategy that minimizes the total cost is termed *computationally-optimal*.

The approach is characterized by explicit modeling of the cost and resource consumption uncertainties inherent in the problem-solving process and methods from Stochastic Dynamic Programming [9]. We construct computationally-optimal K-bound (invoking at most K methods) and ∞-bound on-line control strategies for uninterruptible, independent solution methods. The problem is shown to be NP-complete and the resulting strategies are adaptive to unpredictable external changes in cost and resource availability.

There are $M^K$ possible ways to select $K$ methods out of $M$ independent ones. Using Stochastic Dynamic Programming, we develop a pseudopolynomial-time algorithm consuming $\mathcal{O}(KV^2 \log V)$ space, and running in $\mathcal{O}(KMV^2D^2)$ time, where $D$ is the maximum number of values in cost or resource consumption distributions and $V$ is the largest number between an alternative cost, a limit on resource consumption, and K.

*Supported by Rockwell International under IR&D No. 837. Net address: einav@rpal.com.

[†]Supported in part by a grant from Rockwell International Science Center. Net address: fehling@bayes.stanford.edu.

In some situations there are many applicable solution methods. Some of them may be optimal, others may be of approximate or heuristic nature, and all may have uncertainty in deliberation and execution costs and resource consumption. Subject to resource availability, a number of methods could be sequentially explored in a deliberation phase in order to execute the least costly solution. Problems of this type occur in automatic emergency-response systems, design automation, query optimization, destructive testing, and other areas characterized by significant computational cost or limited deliberation resources. Adapting the response of the system to available resources provides a new approach to real-time systems.

The remainder of the paper is organized as follows: Section 2 presents a motivating example, Section 3 formally states the problem, and in Sections 4 through 6 a family of pseudopolynomial-time algorithms to construct generate-and-test strategies is gradually introduced. Section 4 presents a basic case — K-bound (including at most K steps) strategy with no deliberation cost and no resource consumption; Section 5 — K-bound strategy with deliberation cost and resource consumption; and Section 6 — $\infty$-bound computationally-optimal strategy (no predefined limit on the number of steps). Section 7 discusses related work and Section 8 summarizes our results and presents problems for future research.

## 2 Motivating Example

Consider the following hypothetical situation, due to a leak of some explosive, corrosive gas into a space station's air, a state of emergency is automatically declared. There are two reasons to avoid accumulation of gas – (a) to minimize the damage due to corrosion, and (b) to prevent critical accumulation that can cause an explosion. Since a high concentration of gas can cause an explosion, the response is time-bound.

Let us presume that a number of methods are stored in the station's main computer to deal with various contaminations, utilizing such alternative tactics as isolating contaminated areas, chemical neutralization, and dehermetization of non-vital sectors. The methods differ in their effectiveness, in the amount of damage they cause, and in other material losses; their effects are uncertain and are encoded by probability distribution functions.

Given the details of a specific accident, the estimated effect of any method can be determined by computer simulation. Simulation running times are also uncertain and depend on the inputs and the method that is simulated. We assume that these simulations are computationally intensive and must be performed sequentially, and that due to some random factors the repeated estimation of a method can produce a different estimate but every such estimate corresponds to an executable solution. After evaluating by simulation the effects of several methods, a least costly solution (in terms of execution) will be selected.

The central questions addressed by this paper are: What methods to evaluate? When to stop deliberation and start acting? How can the strategy be adapted in case the external conditions change in the middle of its implementation?

The cleanup operations that must be performed entail a cost of corrective action and should be as low as possible. Since deliberation causes delay in action and increased concentration of the gas due to the leak, the total time of deliberation must be bounded, and may be modeled as a resource constrained.

We assume that the distributions of effects are such that no method stochastically dominates another one (no one is clearly better for all possible outcomes). Our basic observation is that, because of a time bound, it is not possible to estimate all the solutions. The problem is defined precisely in the next section.

## 3 Problem Statement

Let $\mathcal{M} = \{M_1, .., M_M\}$ be a finite set of methods that solve some specific problem-instance $P$. Every method in $M_i \in \mathcal{M}$ computes a solution instance $s_P^{M_i} = M_i(P)$ out of the set $O_i^P$ of the possible solutions. Since we will consider only one problem-instance at a time the index $P$ will be omitted in the following.

We assume that the methods cannot be interrupted. The only exception is when a method exhausts all the available resource, in which case it halts automatically without producing any solution.

Let $Cost(s^{M_i})$ denote the distribution of cost to execute $s^{M_i}$. We denote by $Cost(M_i)$, and $Res(M_i)$ the distributions of cost and resource consumed by method $M_i$ during the deliberation, *i.e.*, computation of $s^{M_i}$. We assume that cost and resource distributions are given as sets of rational probabilities over the finite sets of nonnegative integer values.

In general, a strategy, $S$, will generate and estimate a sequence of $l^S$ methods. The methods in this sequence, and the corresponding solutions obtained by evaluating them are denoted by $M_i^{S\,1}$, and $s^{M_i^S}$, for $i = 1, .., l^S$.

---
[1] Note that in general $M_i^S \neq M_i$

After analyzing a (possibly empty) sequence of the methods evaluated so far and results of these evaluations, a strategy selects a new method to be evaluated next. When the strategy halts, a least costly known solution is executed. If we denote the halting decision by $H$, strategy $S$ can be depicted as performing at every iteration the following generate and test steps (see Figure 1):

$$Generate^S(\{(M_1^S, s^{M_1^S}),..,(M_l^S, s^{M_l^S})\}) =$$

$$= \begin{cases} H & \text{if deliberation halts} \\ M_{l+1}^S & \text{otherwise} \end{cases}$$

$$Test^S(M_{l+1}^S) = s^{M_{l+1}^S}$$

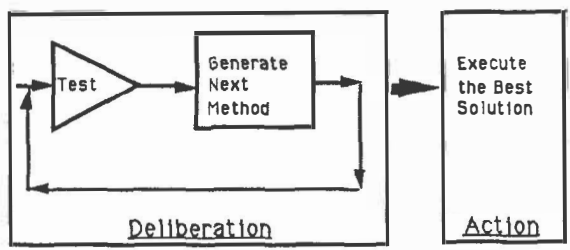

Figure 1: Generate-and-test deliberation strategy.

**Definition 3.1** *$S$ is bounded by $L \in Z^+$ if $l^S \leq L$ for any possible application of $S$. If $S$ is bounded by $L$ we will call it $L$-bound.*

We assume that we are always given a solution $s^{Alt}$ that has cost $Cost^{Alt}$ to be paid if no methods are used at all. For notational convenience we will refer to $Cost^{Alt}$ also as $cost(s^{M_0^S})$. We define:

$$cost(s^S) = \min_{i=0,..,l^S} cost(s^{M_i^S})$$

Using this definition, a least costly known solution will be selected for execution and doing nothing is always a possibility. Let us denote by $Cost(s^S)$ the distribution of $cost(s^S)$ prior to starting strategy $S$, and by $E(Cost(s^S))$ its expected value. Similarly, let $cost(S) = \sum_1^{l^S} cost(M_i^S)$, and $res(S) = \sum_1^{l^S} res(M_i^S)$ denote the total cost and resource consumption in the deliberation phase of $S$.

**Definition 3.2** *Strategy $S$ is feasible if for any possible application of $S$, $res(S) \leq \bar{r}$, where $\bar{r}$ denotes the amount of resource available for deliberation.*

Note that following our earlier definition of uninterruptible methods, any strategy is feasible, since it will simply halt when the deliberation resource is exhausted. Nevertheless, we explicitly specify feasibility to emphasize this point.

Since the exact sequence of methods to be evaluated by $S$ depends on the outcome of the previous evaluations, the deliberation cost is uncertain and described by some random variable $Cost(S)$. Let $E(Cost(S))$ denote its expected value.

**Definition 3.3** *Feasible strategy $S^*$ is computationally-optimal if for any feasible strategy $S$, $E(\text{Cost}(S^*)) + E(\text{Cost}(s^{S^*})) \leq E(\text{Cost}(S)) + E(\text{Cost}(s^S))$.*

In the following we will always mean computational-optimality, when optimality is referred to. We also consider only bounded computationally-optimal strategies fitting our generate-and-test model.

We want to determine whether there is a feasible strategy with expected cost below $\mathcal{C}$. The problem is defined by a 7-tuple — $(\mathcal{C}, \mathcal{M}, Cost^\mathcal{M}, Res^\mathcal{M}, Cost(s^\mathcal{M}), \bar{r}, Cost^{Alt})$ where $\mathcal{M}$ is the set of solution methods, $Cost^\mathcal{M}$ and $Res^\mathcal{M}$ are the distributions of computational costs and resources incurred when the methods are used. $Cost(s^\mathcal{M})$ is the corresponding set of distributions of a solution costs, $\bar{r}$ denotes the amount of resource available for deliberation, and $Cost^{Alt}$ is a finite alternative cost that will be paid, if no methods are used at all.

## 4 K-bound Strategies with No Cost and No Resource Consumption in the Deliberation Phase

In this section we set $cost(M_i) = res(M_i) = 0$, for all $i = 1,.., M$. We begin by stating independence assumptions for the probability distributions of execution cost:

**Assumption 1** *For any $i \neq j$, distributions $\text{Cost}(s^{M_i^S})$ and $\text{Cost}(s^{M_j^S})$ are independent.*

Since nothing prevents a strategy from selecting and estimating the same method more than once, this assumption implies that the cost distributions in any two execution instances of a method are independent.

Assumption 1 is very strong, and it clearly restricts the applicability of the results herein. However, since releasing it significantly complicates the problem, delaying the treatment of interdependent methods is justified for two reasons: one, a simpler algorithm can be obtained when Assumption 1 does hold and two, even when the assumption does not hold, solving a simplified problem can provide a first approximation to a more complex model.

**Definition 4.1** $S^*$ *is a K-bound optimal strategy if $l^{S^*} \leq K$, and for any K-bound feasible strategy $S$, $E(\text{Cost}(s^{S^*})) \leq E(\text{Cost}(s^S))$.*

We will now introduce the table $C^k$ of optimal $k$-step strategies. By conventions of dynamic programming, $C^k$ corresponds to the optimal strategy for the last $k$ methods – $C^1$ corresponding to an optimal 1-step strategy, $C^2$ to an optimal 2-step strategy, and so on.

**Definition 4.2** $C^k(c^{Alt})$ *is the optimal expected cost for all the feasible strategies consisting of exactly $k$ methods and initiated with $c^{Alt}$.*

## 4.1 Computing $K$-bound Optimal Strategy

**Definition 4.3** $C^k(M_i, c^{Alt})$ *is the optimal expected cost for all the feasible strategies consisting of exactly $k$ methods, initiated with $c^{Alt}$, and beginning with method $M_i$.*

We start by calculating $C^1(M_i, c^{Alt})$ for $i = 1, .., M$ and $c^{Alt} = 0, 1, .., Cost^{Alt}$. Although our probability distributions are discrete, the notation drawn from the continuous case makes presentation simpler, and therefore will be used. There is of course a straightforward mapping to the discrete case.

$$C^1(M_i, c^{Alt}) = c^{Alt}[1 - \int_0^{c^{Alt}} dcost(s^{M_i})] +$$
$$+ \int_0^{c^{Alt}} cost(s^{M_i}) dcost(s^{M_i})$$

If method $M_i$ produces a solution with execution cost higher than $c^{Alt}$, the new solution is ignored. The second term represents the contribution to the expected cost when $cost(s^{M_i})$ is lower than $c^{Alt}$. After $C^1(M_i, c^{Alt})$ is computed, we compute

$$C^1(c^{Alt}) = \min_{i=1,..,M} C^1(M_i, c^{Alt})$$

In a general case, ($k = 2, 3, ...; i = 1, .., M; c^{Alt} = 0, 1, .., Cost^{Alt}$),

$$C^k(M_i, c^{Alt}) = C^{k-1}(c^{Alt})[1 - \int_0^{c^{Alt}} dcost(s^{M_i})] +$$
$$+ \int_0^{c^{Alt}} C^{k-1}(cost(s^{M_i})) dcost(s^{M_i})$$
(1)

The argument is similar to the one for $k = 1$. If the cost of computed solution, $cost(s^{M_i})$, is higher than $c^{Alt}$, we keep an old alternative cost, and if the cost is lower, the alternative cost is lowered. In both cases after estimating $M_i$, we have one method less to estimate, therefore $k$ decreases by 1. $C^k(c^{Alt})$ ($k = 2, 3, ...;$ and $c^{Alt} = 0, 1, .., Cost^{Alt}$) is found by:

$$C^k(c^{Alt}) = \min_{i=1,..,M} C^k(M_i, c^{Alt}) \qquad (2)$$

We will denote by $M^k(c^{Alt})$ the method that attains the minimum. If we set $C^0(c^{Alt}) = c^{Alt}$, (1), (2) hold for all $k \in Z^+$. We also define $C^0(M_i, c^{Alt}) = c^{Alt}$.

**Theorem 4.1** $C^k(c^{Alt})$ *defined recursively by (1) and (2) is the optimal expected cost for all the strategies consisting of exactly $k$ methods and initiated with $c^{Alt}$.*

**Proof:** By induction on $k$. $C^0(c^{Alt}) = c^{Alt}$ is the optimal expected cost when no method can be estimated. Assume that $C^{k-1}(c^{Alt})$ is the optimal expected cost for all the strategies consisting of exactly $k - 1$ methods and initiated with $c^{Alt}$. To prove the claim for $C^k(c^{Alt})$, we notice that an optimal strategy must select one of the methods ($M_1, .., M_M$) to be estimated first. Based on our assumption about optimality of $C^{k-1}(c^{Alt})$ and our argument earlier in this section, the expected cost for all the strategies that begin with some method $M_i$ and contain exactly $k$ steps is given by (1), and since an optimal strategy may select the best first method to minimize the expected cost (2) gives the needed optimal expected cost. ∎

To implement the optimal $K$-bound strategy using the table $C^k(c^{Alt})$, we must reconcile a difference in definitions. We defined $K$-bound strategy to include *at most* $K$ methods, while the entries in $C^k(c^{Alt})$ provide an optimal expected values for strategies with *exactly* $k$ methods. It turns out that the two are equivalent. To prove that we will need the following lemma:

**Lemma 4.1** *For any $k \in Z$, $0 \leq a \leq b \Rightarrow C^k(a) \leq C^k(b)$.*

**Proof:** Since $C^k(a) = \min_{i=1,..,M} C^k(M_i, a)$, it will suffice to prove the more specific result: for any $k, i \in Z$, and $0 \leq a \leq b \Rightarrow C^k(M_i, a) \leq C^k(M_i, b)$. By mathematical induction on $k$. For $k = 0$ the claim is true by definition of $C^0(M_i, a)$. Assume it is true for $k - 1$, we must prove that it holds for $k$ as well. Our induction assumption and (2) imply that $C^{k-1}(a) \leq C^{k-1}(b)$. (We will use this fact later.) We must prove:

$$C^k(M_i, a) \leq C^k(M_i, b)$$

or equivalently, using (1),

$$C^{k-1}(a)[1 - \int_0^a dcost(s^{M_i})] +$$

$$+ \int_0^a C^{k-1}(cost(s^{M_i})) dcost(s^{M_i}) \leq$$

$$\leq C^{k-1}(b)[1 - \int_0^b dcost(s^{M_i})] +$$

$$+ \int_0^b C^{k-1}(cost(s^{M_i})) dcost(s^{M_i})$$

or

$$C^{k-1}(a)[\int_a^b dcost(s^{M_i}) + \int_b^\infty dcost(s^{M_i})] +$$

$$+ \int_0^a C^{k-1}(cost(s^{M_i})) dcost(s^{M_i}) \leq$$

$$\leq C^{k-1}(b) \int_b^\infty dcost(s^{M_i}) +$$

$$+ \int_0^a C^{k-1}(cost(s^{M_i})) dcost(s^{M_i}) +$$

$$+ \int_a^b C^{k-1}(cost(s^{M_i})) dcost(s^{M_i})$$

by eliminating identical terms on both sides,

$$C^{k-1}(a)[\int_a^b dcost(s^{M_i}) + \int_b^\infty dcost(s^{M_i})] \leq$$

$$\leq \int_a^b C^{k-1}(cost(s^{M_i})) dcost(s^{M_i}) +$$

$$+ C^{k-1}(b) \int_b^\infty dcost(s^{M_i})$$

This is the sum of two inequalities that follow, as noted, from the induction assumption:

$$C^{k-1}(a) \int_a^b dcost(s^{M_i}) \leq$$

$$\leq \int_a^b C^{k-1}(cost(s^{M_i})) dcost(s^{M_i})$$

and

$$C^{k-1}(a) \int_b^\infty dcost(s^{M_i}) \leq C^{k-1}(b) \int_b^\infty dcost(s^{M_i}) \blacksquare$$

Now the equivalence theorem can be proved.

**Theorem 4.2** *For any $k \in Z$, $C^k(c^{Alt})$ is the expected value of an optimal k-bound strategy initiated with $c^{Alt}$.*

**Proof:** It is enough to prove that for any value of $c^{Alt}$, $l \geq m \Rightarrow C^l(c^{Alt}) \leq C^m(c^{Alt})$. Then every $k$-bound strategy will contain exactly $k$ steps, and therefore $C^k(c^{Alt})$ table could be used. We will prove that for any $c^{Alt}$ and for all $k \in Z$, $C^{k+1}(c^{Alt}) \leq C^k(c^{Alt})$. Indeed, for any $i = 1, .., M$,

$$C^{k+1}(M_i, c^{Alt}) \stackrel{def}{=} C^k(c^{Alt})[1 - \int_0^{c^{Alt}} dcost(s^{M_i})] +$$

$$+ \int_0^{c^{Alt}} C^k(cost(s^{M_i})) dcost(s^{M_i}) \stackrel{Lemma 4.1}{\leq}$$

$$\leq C^k(c^{Alt})[1 - \int_0^{c^{Alt}} dcost(s^{M_i})] +$$

$$+ C^k(c^{Alt}) \int_0^{c^{Alt}} dcost(s^{M_i}) \stackrel{def}{=} C^k(M_i, c^{Alt})$$

and the result follows from the definition of $C$. ∎

By Theorems 4.1 and 4.2 the following Strategy 4.1 is optimal $K$-bound strategy.

**Strategy 4.1.**

Init     Compute $C^k(c^{Alt})$ for $k = 0, 1, 2, .., K$;
          and $c^{Alt} = 0, 1, .., \text{Cost}^{Alt}$.
          Set $c^{Alt} = \text{Cost}^{Alt}$
             $s^* = s^{Alt}$

Step k    For $k = K$ to 1
          Estimate $M^k = M^k(c^{Alt})$
          If $\text{cost}(s^{M^k}) \leq c^{Alt}$
          then Set $c^{Alt} = \text{cost}(s^{M^k})$
                $s^* = s^{M^k}$

**Return $s^*$**

## 4.2 Numerical Example

Consider the following problem: there are only two methods - $M_1$, and $M_2$ ($\mathcal{M} = \{M_1, M_2\}$). The values of the solutions that are produced by these methods can be only 0, 1, or 2 ($D = 3$). We need to compute an optimal strategy for 3 periods ($K = 3$), and we start with an alternative solution of cost 2 (*i.e.*, $\text{Cost}^{Alt} = 2$, it will cost us 2 if we decide to do nothing). Distributions of $cost(s^{M_k})$ for $k = 1, 2$ are shown on Figure 2.

Starting with an optimal strategy for one (last) period ($k = 1$), we compute for $c^{Alt} = 0$, $C^1(M_i, 0)$ for $i = 1, 2$.
$C^1(M_1, 0) = 0$; $C^1(M_2, 0) = 0 \Rightarrow M^1(0) = M_1$ (we recall that $M^1(0)$ denotes a method for which $C^1(M_i, 0)$ is minimized); $C^1(0) = 0$.
For $c^{Alt} = 1$, we obtain
$C^1(M_1, 1) = 0.6$; $C^1(M_2, 1) = 0.5 \Rightarrow M^1(1) = M_2$;
$C^1(1) = 0.5$.

| Values \ Methods | 0 | 1 | 2 |
|---|---|---|---|
| $M_1$ | .4 | .5 | .1 |
| $M_2$ | .5 | .1 | .4 |

Figure 2: Cost distributions of the solution.

For $c^{Alt} = 2$,
$C^1(M_1, 2) = 0.7$; $C^1(M_2, 2) = 0.9 \Rightarrow M^1(2) = M_1$;
$C^1(2) = 0.7$. And so on. The resulting optimal strategy is shown on Figure 3. This strategy

| $c^{Alt}$ \ k | 3 | 2 | 1 | 0 |
|---|---|---|---|---|
| 2 | .153 $M_2$ | .32 $M_1$ | .7 $M_1$ | 2 H |
| 1 | .125 $M_2$ | .25 $M_2$ | .5 $M_2$ | 1 H |
| 0 | 0 $M_1$ | 0 $M_1$ | 0 $M_1$ | 0 H |

Figure 3: Resulting optimal strategy.

will start by estimating method $M^3(2) = M_2$ (column $k = 3$, and row corresponding to $c^{Alt} = 2$). Assume arbitrarily that $cost(s^{M_2}) = 2$, meaning that the alternative cost was not reduced at Step 3. We estimate next method $M^2(2) = M_1$. When (also arbitrarily) we find that $cost(s^{M_1}) = 1$, the alternative cost is reduced to 1 and we continue by estimating $M^1(1) = M_2$. Assume $cost(s^{M_2}) = 2$. At this point deliberation halts, and since the current alternative cost is 1, solution $s^{M_1}$ found at Step 2 is executed.

### 4.3 Computational Complexity
**Input:** M methods are encoded in the input stream. Every method $M_i$, $i = 1, .., M$ is described by the distribution of the cost of the solution, given as a set of rational probabilities over the nonnegative integer values, also given in the input. If the maximum number of values in cost distributions is $D$, and $V = \max\{Cost^{Alt}, K\}$ the length of the input will be $\mathcal{O}(MDlogV)$.
**Space:** Storing the strategy table requires space $\mathcal{O}(KVlogV)$.
**Time:** Computing the table takes time $\mathcal{O}(KMVD)$, since for every entry we must compare $M$ methods, and evaluating each requires $\mathcal{O}(V)$ calculations.

Informally (Garey and Johnson [4]), an algorithm has a polynomial-time complexity if it runs in a time polynomial in the length of its input. $KMVD$ is not a polynomial function of $MDlogV$, but it is a polynomial function of $MDlogV$, $K$, and $V$. An algorithm that runs in time polynomial in its input length and the largest number in the input has a pseudopolynomial-time complexity, which is true in our case.

**Conjecture 4.1** *Computing computationally-optimal real-resource strategy with no cost and no resource consumption in deliberation phase is NP-complete if $D \geq 3$.*

In the next section, we prove the NP-completeness of a more general problem.

## 5 K-bound Strategies with Cost and Resource Consumption in the Deliberation Phase

In this section we extend the basic technique developed in Section 4 to handle cost and resource consumption in the deliberation phase. First we introduce the cost alone. Two new assumptions are needed.

**Assumption 2** *For any $i \neq j$, distributions $\text{Cost}(M_i^S)$ and $\text{Cost}(M_j^S)$ are independent.*

**Assumption 3** *For any $i$ and $j$, distributions $\text{Cost}(M_i^S)$ and $\text{Cost}(s^{M_j^S})$ are independent.*

### 5.1 Cost in the Deliberation Phase, No Resource Consumption
Presence of the cost in deliberation phase may cause an optimal strategy to estimate fewer methods. We will appropriately change the definitions of $C^k(c^{Alt})$ and $C^k(M_i, c^{Alt})$:

**Definition 5.1** $C^k(c^{Alt})$ *is the optimal expected total cost for all the strategies consisting of at most $k$ methods and initiated with $c^{Alt}$.*

**Definition 5.2** $C^k(M_i, c^{Alt})$ *is the optimal expected total cost for all the strategies consisting of at most $k$ methods, initiated with $c^{Alt}$, and beginning with method $M_i$.*

The recursive formula for $C^k(M_i, c^{Alt})$ will be changed to reflect the cost of deliberation:

$$C^k(M_i, c^{Alt}) = C^{k-1}(c^{Alt})[1 - \int_0^{c^{Alt}} dcost(s^{M_i})] + \quad (3)$$

$$+ \int_0^{c^{Alt}} C^{k-1}(cost(s^{M_i})) dcost(s^{M_i}) + E(Cost(M_i))$$

Note that by Assumption 3 only the expected value of deliberation cost appears in this formula.

Finally, to provide for possibility that in some situations doing nothing could be the best strategy, we introduce a new artificial method to be denoted by $M_0$. It is characterized by $cost(M_0) = 0$ and $cost(s^{M_0}) = Cost^{Alt}$. Since the deliberation cost of $M_0$ is 0 it can be used without restriction and since $cost(s^{M_0}) = Cost^{Alt}$ it will never improve a current alternative solution.

The definition of $C^k(c^{Alt})$ will be altered to include $M_0$:

$$C^k(c^{Alt}) = \min_{i=0,1,..,M} C^k(M_i, c^{Alt}) \quad (4)$$

**Theorem 5.1** $C^k(c^{Alt})$ *computed recursively by (3) and (4) is the optimal expected total cost for all the strategies consisting of at most k methods with cost in deliberation phase and initiated with* $c^{Alt}$.

**Proof:** Proof is similar to that of Theorem 4.1. We must only notice that since the deliberation cost is independent of execution cost, the expected deliberation cost is included in (3) and since the dummy method $M_0$ may now be selected, (4) must include $M_0$. ∎

By Theorem 5.1, if Strategy 4.1 will compute $C^k(c^{Alt})$ using (3) and (4) it will be an optimal strategy with deliberation cost.

The space and computational complexity with deliberation cost are the same as without them, as presented in Section 4.

### 5.2 Both Cost and Resource Consumption in the Deliberation Phase

In this model the estimate of every method causes the consumption of some quantity of a single resource, described by the distribution $Res(M_i)$, $i = 1, 2, .., M$, from the total level – $\bar{r}$ – available in the beginning of the process.

With one exception, we assume uninterruptability of methods: when the total resource consumption reaches $\bar{r}$ the deliberation process is interrupted and the best solution available at this point is executed.

It is assumed that the interrupted method produces no solution and incurs full computational cost.

In order to account for the resource consumption, we must add a resource dimension to the strategy table. Other than that, our discussion parallels the development of the previous models. As with the cost, we make an independence assumptions for resource consumption distribution functions.

**Assumption 4** *For any* $i \neq j$, *distributions* $Res(M_i^S)$ *and* $Res(M_j^S)$ *are independent.*

**Assumption 5** *For any* $i$ *and* $j$, *distributions* $Res(M_i^S)$ *and* $Cost(s^{M_j^S})$; *and* $Res(M_i^S)$ *and* $Cost(M_j^S)$ *are independent.*

Similarly to the previous case, we define:

**Definition 5.3** $C^k(c^{Alt}, r)$ *is the optimal total expected cost for all the strategies consisting of at most k methods, initiated with* $c^{Alt}$, *and that can be executed within resource limit r.*

**Definition 5.4** $C^k(M_i, c^{Alt}, r)$ *is the optimal total expected cost for all the strategies consisting of at most k methods, initiated with* $c^{Alt}$, *that can be executed within resource limit r, and begin with method* $M_i$.

We also define:

$$C^k(c^{Alt}, r) = \min_{i=0,..,M} C^k(M_i, c^{Alt}, r) \quad (5)$$

Where

$$C^k(M_i, c^{Alt}, r) = \int_0^r C^{k-1}(c^{Alt}, r - res(M_i)) dres(M_i) \cdot \quad (6)$$

$$\cdot [1 - \int_0^{c^{Alt}} dcost(s^{M_i}) \int_0^r dres(M_i)] +$$

$$\int_0^r dres(M_i) \int_0^{c^{Alt}} C^{k-1}(cost(s^{M_i}), r - res(M_i)) dcost(s^{M_i})$$

$$+ E(Cost(M_i))$$

**Theorem 5.2** $C^k(c^{Alt})$ *computed recursively by (6) and (5) is the optimal expected cost for all the strategies consisting of at most k methods with cost and resource consumption in deliberation phase, and initiated with* $c^{Alt}$.

**Proof:** The proof is similar to that of Theorem 5.1. We must only notice that since the resource consumption is independent from cost, (6) is correct. ∎

It remains to modify Strategy 4.1 to include resource management. The optimality of the Strategy 5.1 follows from the Theorem 5.2.

**Strategy 5.1.**

Init       Compute $C^k(c^{Alt}, r)$ for $k = 0, 1, 2, .., K$;
              $c^{Alt} = 1, .., \text{Cost}^{Alt}$; and $r = 0, 1, .., \bar{r}$
              Set $c^{Alt} = \text{Cost}^{Alt}$
                  $r = \bar{r}$
                  $s^* = s^{Alt}$

Step k    For $k = K$ to 1
              Estimate $M^k = M^k(c^{Alt}, r)$
              Set $r = r - \text{res}(M^k)$
              If $\text{cost}(s^{M^k}) \leq c^{Alt}$
              then Set $c^{Alt} = \text{cost}(s^{M^k})$
                      $s^* = s^{M^k}$

Return $s^*$

### 5.3 Computational Complexity

**Input:** The length of input is $\mathcal{O}(MD\log V)$, where, as before, $D$ is the maximum number of values in cost or resource consumption distributions, and $V = \max\{Cost^{Alt}, K, \bar{r}\}$.

**Space:** Storing the strategy table requires space $\mathcal{O}(KV^2 \log V)$.

**Time:** Computing the table takes time $\mathcal{O}(KMV^2D^2)$, because for every entry we must compare $M$ methods, and evaluating each requires $V^2$ calculations. The algorithm has pseudopolynomial-time complexity.

**Theorem 5.3** *Computing computationally-optimal real-resource strategy is NP-complete problem.*

**Proof Outline:** Our problem can be easily solved by nondeterministic automaton by branching nondeterministically every time a new method must be generated. Next, we will show that our problem restricts to Integer Knapsack Problem (see p. 247 [4]) *i.e.*, Integer Knapsack Problem is a special case of our problem. Set $C = K$. Define a method for each $u \in U$: set $cost(M_i) = 0$, $res(M_i) = s(u)$, and define the distributions of execution cost by $p(cost(s^{M_i}) = 1) = 10^{-v(u)}$, $p(cost(s^{M_i}) = 0) = 1 - 10^{-v(u)}$. Set $\bar{r} = B$, $Cost^{Alt} = 1$. Solving this real-resource strategy problem will solve Integer Knapsack Problem. ∎

## 6 Strategies without Predefined Step Limit

We now consider the most general case, corresponding to our example in Section 2. In real-life situations we do not typically restrict the number of steps, or iterations to be taken by a strategy. Our concern is that the strategy be optimal while the exact number of steps is not important. We are still insisting that the strategy be bounded; otherwise it may never terminate, but we may not be given a specific bound.

This section shows that in some cases — namely when all methods have positive deliberation cost — a problem with unspecified bound may be reduced to $K$-bound problem by calculating the upper bound. In other cases the $\infty$-bound problems may have no bounded, optimal strategies — resource permitting, it will always be beneficial to add another method in hope to reduce the execution cost even further.

Theorem 6.1 finds the bound for problems with positive deliberation cost.

**Theorem 6.1** *For any instance of a problem with positive deliberation cost. If $C_{min} = \min_{i=1,..,M} cost(M_i)$ an $\infty$-bound optimal strategy is bounded and equivalent to an optimal strategy for $K'$-bound problem, where $K' = \lceil \frac{Cost^{Alt}}{C_{min}} \rceil$.*

**Proof:** Let $P$ be an instance of a problem with positive deliberation cost. The deliberation cost of any strategy for $P$ using more than $K'$ steps will exceed the alternative cost, so such a strategy cannot be optimal. Since the number of the possible strategies containing at most $K'$ steps is finite, an optimal strategy exists, and has at most $K'$ steps. ∎

The implementation of $\infty$-bound strategy is obvious: 1) compute $K'$; 2) implement $K'$-bound optimal strategy.

## 7 Related Work

Our work emphasizes the analysis of computationally-optimal control of deliberation before action. Several approaches were suggested to related problems involving the control of deliberation.

Tokawa and Kim [10] treat a similar component-selection problem in design-automation domain. They suggest selecting components in random, and evaluate the resulting strategy by its rank in a list of all possible strategies. Ono and Lohman [7], consider the problem of query optimization. They construct a polynomial strategy for some types of queries using a dynamic programming formulation, but they do not allow for uncertainty, and do not consider a resource constraint.

Brooks [2], and Agre and Chapman [1] suggest reactive approach to deliberation problem, and although Brooks provides for deliberation when reaction fails, neither Brooks', nor Agre and Chapman's work offers optimal deliberation control methods as those presented here.

Kaelbling [6] and Rosenshein and Kaelbling [8] suggest avoiding deliberation by compiling in ad-vance the programs responding within the given resource limit, but they do not allow for uncertainty, and do not claim optimality with respect to all the programs meeting the limit.

Horvitz [5] offers a decision-theoretic algorithm to select a single method among the set of alternative interruptible methods, and Fehling and Breese [3] present a computational architecture and decision-theoretic principles for real-time control.

In general, the reactive approaches focus on eliminating the deliberation by sacrificing the optimality, while the decision-theoretic work is traditionally ignoring the computational issues in pursued of an optimal solution. By introducing the *computationally-optimal* strategies we reach an optimal balance between the deliberation and execution costs.

## 8 Summary and Future Work

In this paper we have stated a problem of finding *computationally-optimal* real-resource strategies for independent, uninterruptible solution methods, and shown how to solve it for all practical purposes. The problem, which we have shown to be NP-complete, appears in numerous practical applications. We developed an algorithm that solves it in a polynomial time if the alternative cost, resource limit, and number of steps have small values.

The results can be readily extended to the case of multiple resources. This extension involves adding a dimension to the strategy table for each new resource, it does not require any new technical ideas and is left to the reader.

Future work may address validating our NP-completeness conjecture, allowing dependencies among the methods, and considering interruptible methods.

## 9 Acknowledgments

We would like to thank Michael Genesereth, Matt Ginsberg, Eric Horvitz, Ross Shachter, and Dave Smith for helpful discussions that resulted in improved presentation of this paper. Diane Cunliffe and Sue Kenney helped greatly to improve the style. We also thank anonymous referees for their suggestions. The first author is grateful to Palo Alto Laboratory of the Rockwell International Science Center for creating an ideal environment for this work.